\RequirePackage{fix-cm}
\documentclass[smallextended]{svjour3}       
\smartqed  
\usepackage{graphicx}
\usepackage{natbib}
\usepackage{todonotes}
\usepackage{comment}
\usepackage{multirow}
\usepackage{amssymb}
\usepackage{amsmath}
\usepackage{algorithm}
\usepackage{algorithmic}
\usepackage{setspace}
\usepackage[hidelinks]{hyperref}

\newtheorem{myDef}{Definition}
\newcommand{\SET}[1]{\mathcal{#1}}

\newcommand{\s}[1]{\mathit{#1}}

\newif\ifcomment \commentfalse 
\begin{document}

\title{Explainable AI Enabled Inspection of Business Process Prediction Models}

\titlerunning{XAI Enabled Inspection of Business Process Prediction Models}        

\author{Chun~Ouyang \and Renuka~Sindhgatta \and Catarina~Moreira}


\institute{C. Ouyang, R. Sindhgatta, C. Moreira \at
              Queensland University of Technology, Brisbane, Australia \\
              \email{\{c.ouyang,renuka.sr,catarina.pintomoreira\}@qut.edu.au}           
}

\date{Received: date / Accepted: date}

\maketitle

\begin{abstract}
Modern data analytics underpinned by machine learning techniques has become a key enabler to the automation of data-led decision making. 
As an important branch of state-of-the-art data analytics, business process predictions are also faced with a challenge in regard to the lack of explanation to the reasoning and decision by the underlying `black-box' prediction models. 
With the development of interpretable machine learning techniques, explanations can be generated for a black-box model, making it possible for (human) users to access the reasoning behind machine learned predictions. 
In this paper, we aim to present an approach that allows us to use model explanations to investigate certain reasoning applied by machine learned predictions and detect potential issues with the underlying methods thus enhancing trust in business process prediction models. 
A novel contribution of our approach is the proposal of model inspection that leverages both the explanations generated by interpretable machine learning mechanisms and the contextual or domain knowledge extracted from event logs that record historical process execution. 
Findings drawn from this work are expected to serve as a key input to developing model reliability metrics and evaluation in the context of business process predictions. 
\end{abstract}


\section{Introduction}
\label{sec:intro}

Business processes form a lifeline of business within and across organisations. Executions of day-to-day business processes involve a wide range of stakeholders and are supported by a variety of information systems. Data generated by process executions are stored in event logs. In recent years, machine learning techniques are being applied to construct \textit{business process prediction models}. These models aim at predicting future states of a business process by learning from process execution history recorded in event log data. Typical examples of business process predictions include predicting outcomes that the execution of a business process may lead to~\citep{teinemaa2019} and the remaining time till the completion of process execution~\citep{verenich2019}. 


However, similarly to other state-of-the-art data analytics, business process predictions are also faced with a challenge in regard to the lack of explanation to the reasoning and decision of its prediction models. While existing studies focus on applying advanced, complex learning techniques that can achieve high accuracy of process predictions, these techniques often have sophisticated internal computations that are largely recalcitrant to explanation and thus are often applied and recognised as a `black-box'. 

The recent body of literature in machine learning has emphasised the need to understand and trust the predictions (e.g.,~\cite{lakkaraju2019,rudin2019}), and methods and techniques have been proposed for explaining black-box models. These are known as \textit{interpretable machine learning} or, in a broader context, \textit{explainable AI} (XAI). In particular, explanations from a black-box model make it possible for (human) users to confirm multiple prerequisites of a machine learning model~\citep{DoshiVelez2017}, which are fairness, reliability and robustness, causality, privacy and trust. In their widely-cited survey of XAI methods, \citet{guidotti2018} point out that the ``applications in which black box decision systems can be used are various, and each approach is typically developed to provide a solution for a specific problem''. 

In this paper, we aim to present an approach that allows us to use model explanations to investigate certain reasoning applied by machine learned predictions and detect potential issues with the underlying methods thus enhancing trust in business process prediction models. On the one hand, it is challenging to generate explanations of a black-box model which takes as input the event log data that are complex and capture multi-dimensional (temporal) sequences of process execution. 
On the other hand, event logs often record rich information about process execution, and by applying the existing process mining techniques~\citep{vanderaalst2016process} relevant process knowledge can be extracted from event logs, which provide contextual or domain knowledge useful for understanding model explanations. A novel contribution of our approach is the proposal of model inspection that leverages both the explanations generated by technical interpretable machine learning mechanisms and the contextual or domain knowledge extracted from event logs. Findings drawn from this work are expected to serve as a key input to developing model reliability metrics and evaluation in the context of business process predictions. 


Next, Section~\ref{sec:background} provides background information with a review of related work. 
Section~\ref{sec:approach} proposes our approach to support inspection of business process prediction models. 
Section~\ref{sec:reviewbenchmarks} presents model inspection and analysis in the context of two existing predictive process monitoring benchmarks and using three real-life event logs. 
Finally, Section~\ref{sec:concl} concludes the paper. 


\section{Background and Related Work} 
\label{sec:background}

\subsection{Business Process Event Log}
\label{subsec:eventlog}

\begin{figure}[h!]
    \includegraphics[width=0.9\textwidth]{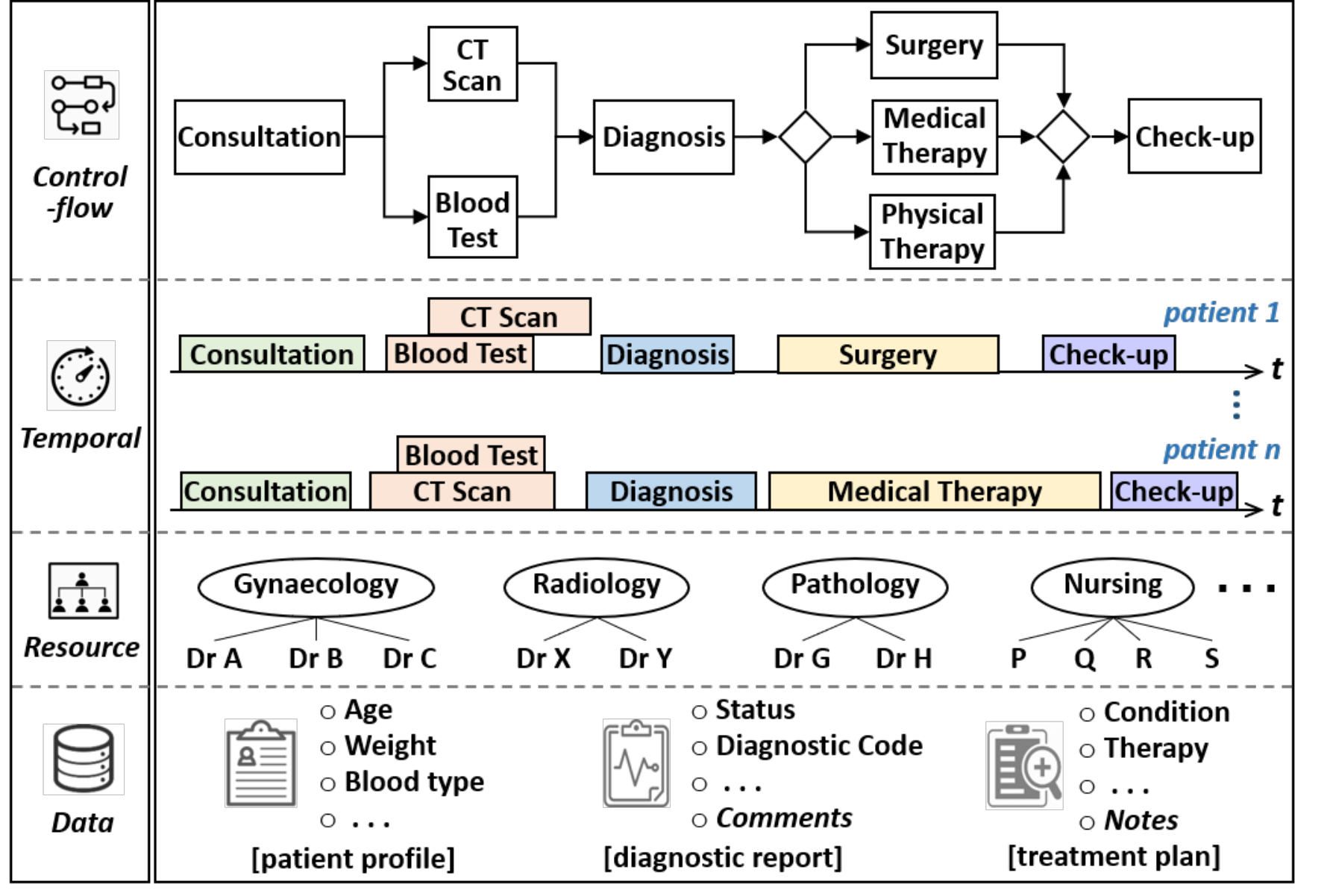}
    \caption{A simplified hospital process and multiple perspectives of process execution}
\label{fig:multiperspective}
\end{figure}

\begin{figure}[h!]
    \includegraphics[width=0.95\textwidth]{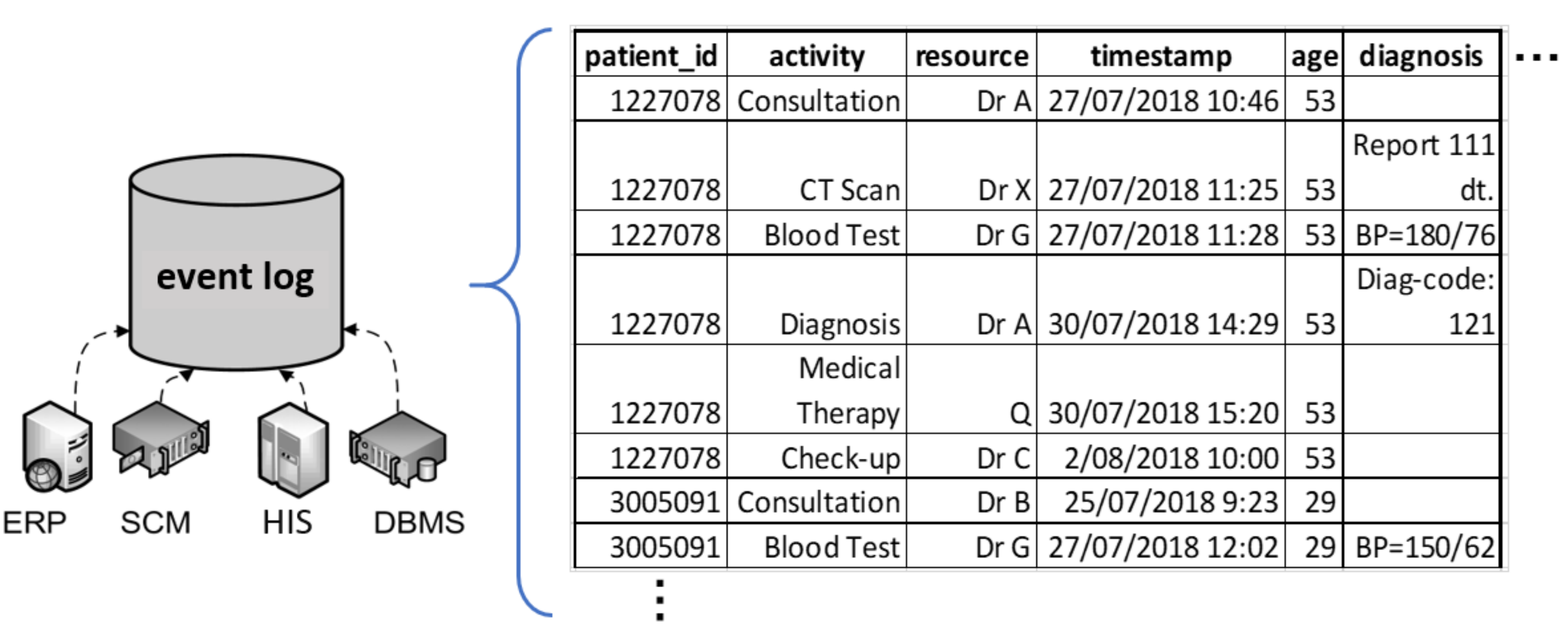}
    \caption{A fragment of an event log that records execution of the process in Fig.~\ref{fig:multiperspective}}
\label{fig:eventlog}
\end{figure}

Business process execution involves multiple perspectives and data generated along process execution are recorded in \textit{event logs}. Fig.~\ref{fig:multiperspective} depicts an example of a simplified hospital process and its multiple execution perspectives, and Fig.~\ref{fig:eventlog} shows a fragment of an event log that records information about the process execution. The example is inspired by the information captured by a real-life event log recorded by the information systems in a Dutch academic hospital\footnote{Refer to Business Process Intelligence Challenge (BPIC) 2011 at \url{https://www.win.tue.nl/bpi/doku.php?id=2011:challenge}.}. The process consists of several medical activities: consultation, test, diagnosis, treatment and check-up. Each instance of process execution (\textit{a.k.a.}~case) captures a hospital episode of a specific patient and can be identified by the patient reference number recorded in the hospital's IT systems.  

An event log comprises \textit{events} and \textit{event attributes} (see Definitions~\ref{def:eventlog} and~\ref{def:eventatt}). Examples of events and event attributes are shown respectively as rows and columns in Fig.~\ref{fig:eventlog}. An event log has a multi-dimensional data structure capturing multiple perspectives of process execution. It consists of: 
(1) \textit{control-flow} data (describing the activities and their occurrences); 
(2) \textit{temporal} data (representing an ordered sequence of timestamps relevant to activity occurrences); 
(3) \textit{resource} data (which is tabular data recording the resources and their attributes and may be extracted from HR databases); and 
(4) \textit{case} data (which is tabular data recording the process instances and their attributes, e.g., patient age group, blood type, disease associated with a hospital episode). 
In addition, an event log may also contain event attributes that record textual data (corresponding to such as notes or comments relevant to process execution).  

\begin{myDef}[Event log]
\label{def:eventlog}
\upshape 
    Let $\SET{E}$ be the set of all event identifiers (or \textit{events} in short), $\SET{A}$ the set of all attribute names, $\SET{V}$ the set of all attribute values.
    An \textit{event log} can be defined as $\s{L}=(\s{E},\s{A},\gamma)$, where $\s{E}\subseteq\SET{E}$ is the set of events, $\s{A}\subseteq\SET{A}$ is the set of event attributes, and $\gamma\in\s{E}\rightarrow(\s{A}\nrightarrow\SET{V})$ specifies event attribute value bindings. 
    Each event $e\in\s{E}$ has event attributes $\s{dom}(\gamma(e))$. 
    For each $a\in\s{dom}(\gamma(e))$, $\gamma_a(e)=\gamma(e)(a)$ is the value of attribute $a$ for event~$e$. 
\end{myDef}

\begin{myDef}[Event attribute]
\label{def:eventatt}
\upshape 
Let $\s{Case}\subseteq\SET{V}$ be the set of case identifiers, $\s{Act}\subseteq\SET{V}$ the set of activity names, $\s{Time}\subseteq\SET{V}$ the set of timestamps, $\s{Res}\subseteq\SET{V}$ the set of resource identifiers, and $\s{Data}\subseteq\SET{V}$ the set of case data values. 
An event log $\s{L}=(\s{E},\s{A},\gamma)$ has three \textit{mandatory} event attributes $\{\s{case},\s{act},\s{time}\}\subseteq\s{dom}(\gamma(e))$ and an arbitrary finite number of additional event attributes $\{\s{res},\s{data}_1,...,\s{data}_n\}\subseteq\s{A}$ (where $n\in\mathbb{N}$). 
For each $e\in E$: 
\begin{itemize}
		\item $\gamma_\s{case}(e)\in\s{Case}$ is the case to which event~$e$ belongs,
		\item $\gamma_\s{act}(e)\in\s{Act}$ is the activity that $e$ refers to,
		\item $\gamma_\s{time}(e)\in\s{Time}$ is the time at which~$e$ occurred, 
		\item $\gamma_\s{res}(e)\in\s{Res}$ is the resource that executed~$e$ (if $\s{res}\in\s{dom}(\gamma(e))$), and 
        \item $\gamma_{\s{data}_i}(e)\in\s{Data}$ (where $i \leq 1 \leq n$) is some case data associated with~$e$ (if $\s{data}_i\in\s{dom}(\gamma(e))$). 
\end{itemize}
\end{myDef}

\subsection{Business Process Prediction}
\label{subsec:benchmarks}

Business process event logs comprising historical process execution are widely used for predicting how a business case, such as a loan application, will unfold until its completion at run time. Such predictions intend to provide better service to customers by mitigating delays or unexpected outcomes. There has been a growing interest in building complex machine learning models for predicting business process behaviour, such as the next event in a case, the time for completion of an event, and the remaining execution trace of a case. Early work focuses on extracting features from business process event logs~\citep{Leontjeva2015}. As most machine learning algorithms need to be configured with certain hyperparameters for better performance based on the characteristics of the event log, a framework for tuning the hyperparameter using genetic algorithms has been proposed~\citep{Chiara2018}. 
A systematic review and taxonomy of methods predicting the business process outcome has been presented with a comparative experimental evaluation of eleven representative methods on nine real-world event logs~\citep{teinemaa2019}. Another study presents a systematic review and taxonomy for predicting the remaining time of a business process instance and provides an experimental evaluation of representative methods on sixteen real-world datasets~\citep{verenich2019}. 

In this work, we focus on the above two benchmark studies by \citet{teinemaa2019} and \citet{verenich2019} that evaluate various techniques used in the context of predictive process monitoring. Existing methods on predicting future behaviour of business process mainly apply supervised machine learning algorithms. A supervised learning algorithm uses a corpus of input, output pairs (or training data) to learn a hypothesis that can predict the output for a new or unseen input (test data). The input and output are derived from the event logs. The output in the two benchmarks are: an outcome of a case defined by using a labelling function, and the remaining time for a case to complete. In both benchmark studies, evaluation of different business process prediction techniques focuses on model accuracy. For outcome-oriented prediction, this is measured by the \textit{area under the ROC curve} (AUC) metric, and for remaining time prediction, this is measured by the \textit{mean absolute error} (MAE) metric. Higher the AUC, better the model is at predicting the outcomes. MAE for remaining time prediction is the absolute difference between the actual remaining time and predicted remaining time and a lower MAE indicates better performance of the prediction model. 

\subsection{Black Box Explanation}
\label{subsec:predictivemodels}

In recent years, the topic of \textit{explainable} and \textit{interpretable machine learning}~\citep{guidotti2018} has gained more and more attention. 
Use of machine learning in critical decision-making processes has urged the need to provide the reasoning that leads to predictions made by these algorithms. Model interpretability can be addressed by having \textit{interpretable models} or providing \textit{post hoc interpretations}. An \textit{interpretable model}~\citep{lipton2018}, is able to provide transparency at the level of entire model, the level of individual components, and the level of the learning algorithm. For example, both linear regression models and decision tree models are interpretable models, while neural network models are complex and hard to follow and hence have low transparency.

Another distinct approach to address model interpretability is via \textit{post hoc interpretation}. Here, explanations and visualisations are extracted from a learned model, that is, \textit{after} the model has been trained, and hence are \textit{model agnostic}. Existing techniques can be divided into two categories: partial dependence models and surrogate models. 
More specially, \textit{surrogate models} use the input data and a black box model (i.e., a trained machine learning model) and emulate the black box model. In other words, they are approximation models that use interpretable models 
to approximate the predictions of a black box model, enabling a decision-maker to draw conclusions and interpretations about the black box~\citep{Molnar18}. 
Interpretable machine learning algorithms, such as linear regression and decision trees are used to learn a function using the predictions of the black box model. 
This means that this regression or decision tree will learn both well classified examples and misclassified ones. 
Measures such as mean squared error are used to assess how close the predictions of the surrogate model approximate the black box.
As a result, the explanations derived from the surrogate model reflect a local and linear representation of the black box model. 

Explanations from a black-box model permits us to confirm multiple prerequisites of a machine learning model~\citep{DoshiVelez2017}, which are \emph{fairness}, \emph{reliability and robustness}, \emph{causality}, \emph{privacy} and \emph{trust}. In this work, we aim to present an approach that allows us to use model explanations to investigate certain reasoning applied by machine learned predictions and detect potential issues thus enhancing \emph{trust} in business process prediction models.

\section{Approach}
\label{sec:approach}

\begin{figure}[b!]
    \includegraphics[width=\textwidth]{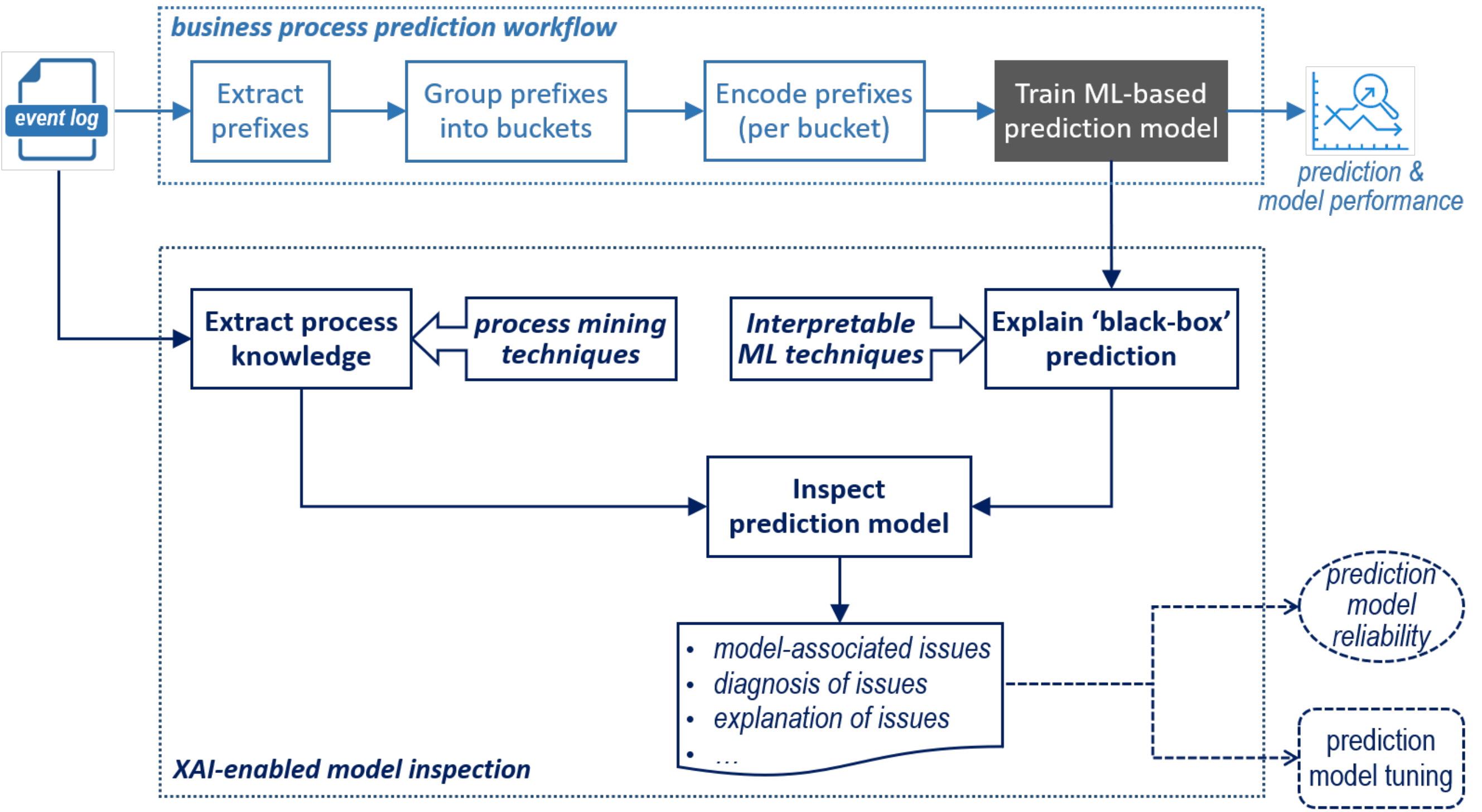}
    \caption{Approach for XAI-enabled inspection of business process prediction models}
\label{fig:approach}
\end{figure}

In this section, we propose an approach to support XAI-enabled inspection of business process prediction models. Fig.~\ref{fig:approach} depicts an overview of the approach. 

A standard workflow for building business process prediction models based on machine learning~\citep{verenich2019} is illustrated at the top of Fig.~\ref{fig:approach}. 
At first, \textit{prefixes} are extracted for each trace (denoted as $\sigma$). $\sigma$ is an ordered sequence of $n$ events $\langle e_1,...,e_n \rangle$ and corresponds to a set of $n$ prefixes $\{\langle e_1,...,e_i \rangle~|~1\leq i\leq n\}$. 
We use $\mathit{prefix}(\sigma,i)$ to denote the sequence of the first~$i$ events extracted for~$\sigma$ and thus $i$ specifies the prefix length. 
Next, the prefixes extracted for all instances of process execution are grouped into \textit{buckets} based on their similarities (such as prefix length or state of process execution) and the prefixes in each bucket are then encoded into feature vectors. 
Different techniques can be applied for bucketing and encoding prefixes and will be discussed in Section~\ref{subsec:design}. 
The buckets of feature vectors are ready as input for a machine learning (ML) algorithm. 
Since the future state for each trace is known from event log data, pairs of the encoded prefix and the future state are used to train a ML-based prediction model for each bucket. 
The resulting prediction model is assessed for its performance using conventional metrics in machine learning (such as accuracy, precision, recall, F1 score). 


Since business process prediction models are mostly built upon advanced machine learning algorithms to achieve better model performance, they become black-box models. 
A novel contribution of our approach is the proposal of process prediction model inspection as depicted in Fig.~\ref{fig:approach}. 
A key step is to generate explanations about the process predictions made by black-box models using interpretable ML techniques. For example, it is important to understand the reasons for machine learned process predictions~\citep{guidotti2018}. Often the output produced by a technical interpretation mechanism requires validation with a human expert or via applying relevant contextual or domain knowledge before it can be used to derive meaningful insights. 
In the field of business process studies, event logs record process execution history and by using process mining techniques~\citep{vanderaalst2016process} various process knowledge can be extracted from event log data, which can serve as contextual or domain knowledge. A typical example is discovery of the (logged) execution paths of process activities (i.e., control-flow of process execution) in the form of a process graph. Hence, we propose the inclusion of another important step in our approach, which focuses on extracting process knowledge relevant to explanations of process prediction. Both the technical interpretation of black-box predictions and relevant process knowledge extracted from event log data are used for inspection of process prediction models. The expected findings from such model inspection include detection of potential issues associated with a process prediction model, diagnosis of the issues, explanations about the issues and/or reasons that may have caused the issues, etc. 

We consider that detection of model-associated issues will serve as a key input to developing model reliability metrics and evaluation, and diagnosis and explanations of the issues will help with model tuning for improving the reliability and robustness of process prediction models in our future work.  



\section{Model Inspection and Analysis}
\label{sec:reviewbenchmarks}

Following the approach proposed in the previous section, we carry out inspection of business process prediction models used for predicting process outcome and remaining time, and analyse the inspection results to draw findings about the prediction models. The source code of model inspection and analysis is available \url{https://git.io/Je186} (regarding process outcome prediction) and \url{https://git.io/Je1XZ} (regarding process remaining time prediction).

\subsection{Design and Configuration}
\label{subsec:design} 

In the context of two existing predictive process monitoring benchmarks (see Section~\ref{subsec:benchmarks}), combinations of various bucketing, encoding, and supervised learning algorithms have been evaluated for predicting process outcome~\citep{teinemaa2019} and remaining time~\citep{verenich2019}, respectively. 
In this study, we decide to choose the following techniques, given the fact that the methods built upon a combination of these techniques have better performance (i.e., high AUC values or low MAE values) as compared to others according to the benchmark evaluation. 

\paragraph{Bucketing techniques:}  
    i) \textbf{single bucket}, where all prefixes of traces are considered in a single bucket, and a single prediction model is trained; and 
    ii) \textbf{prefix length bucket}, where each bucket contains the prefixes of a specific length, and one prediction model is trained for each possible prefix length.
							
\paragraph{Encoding methods:}  
	i) \textbf{static encoding}, where the case attributes of each event that remain the same through out the prefixes of a trace is added as a feature \textit{as-is}. The categorical attributes are one-hot encoded, where each categorical attribute value is represented as a feature that takes a binary value of 0 or 1; and  
	ii) \textbf{aggregation encoding}, where the prefixes in each bucket are transformed by considering only the frequencies of event attributes (such as activity, resource) and computing four features to capture the numeric event attributes (maximum, mean, sum and standard deviation), and note that this way the order of the events in a prefix is ignored.  
	
%

\paragraph{Machine learning algorithms:} 
    We choose \textit{gradient boosted trees}~\citep{friedman2001} (specifically \textbf{XGBoost}), which is used in both benchmarks and outperformed the other machine learning techniques (e.g., random forest, support vector machines, logistic regression). 
    Note that the benchmark study by \citet{verenich2019} considers \textit{Long Short Term Memory} (LSTM) as a deep learning algorithm used for remaining time prediction. Given that deep learning-based models require different feature encoding from traditional machine learning algorithms, LSTM is yet not considered in this study. 
    
\paragraph{Interpretable ML techniques:} 
    To make sure that the output and performance of a business process prediction model under inspection remain intact, we apply \textit{post hoc interpretation} to derive explanations for a process prediction model built upon the above design and configuration. 
    
    When generating explanations of process prediction models, we propose that \textbf{global explanations} can be derived as the first step to gain an overall understanding of a model. These explanations may already help reveal certain issues associated the model. In this study, we conduct \textit{permutation feature importance measurement} supported by gradient boosted trees to obtain relevant global explanations about a process prediction model. More specifically, the feature importance value generated by XGBoost is used to explain the impact of different features to the overall predictions made by a prediction model. 

    Next, we are interested in deriving \textbf{local explanations} about specific process predictions. In this study, we choose a representative local surrogate technique, known as \textit{Local Interpretable Model-Agnostic Explanations} (\textbf{LIME})~\citep{Ribeiro16}, which can explain the predictions of any classification or regression algorithm by approximating it locally with a linear interpretable model. We use LIME to generate local explanations useful in interpreting the prediction for a particular trace of process execution. The explanations by LIME are represented in terms of feature attribution, i.e., the features that have impacted a specific model learned prediction and the value of feature attribution measuring such impact. 
    
    
\paragraph{Process mining platform:} 
    We leverage existing process mining techniques for discovery of process knowledge. 
    The discovered control-flow of process execution is often specified in process graphs. Statistical/descriptive analysis can be applied in the context of process execution. 
    A variety of process mining techniques are supported by existing process mining software platforms. We choose DISCO\footnote{A commercial tool supporting process mining for professionals and providing academic licenses free of charge (\url{https://fluxicon.com/disco/})} as the process mining platform in our experiments.

\subsection{Datasets and Notations}
\label{subsec:data} 

We present model inspection and analysis on three real-life event logs that are representative of our experiments and produce interesting results and findings. The event logs, known as 
BPIC~2011\footnote{\url{https://data.4tu.nl/articles/dataset/Real-life_event_logs_-_Hospital_log/12716513/1}}, BPIC~2012\footnote{\url{https://data.4tu.nl/articles/dataset/BPI_Challenge_2012/12689204}} and BPIC~2015\footnote{\url{https://data.4tu.nl/collections/BPI_Challenge_2015/5065424}}, are from the Business Process Intelligence Challenge initiative and publicly available at 4TU.ResearchData. They were used for evaluation of business process predictions in the two benchmark studies~\citep{teinemaa2019,verenich2019}. Below are brief descriptions of these three event logs as well as certain notations to be used in the experiments. A summary of statistics of the three event logs used for model inspection and analysis is provided in Table~\ref{tab:my-table}. 

\paragraph{BPIC 2011:} The event log contains cases of patient visits to the Gynaecology department of a Dutch hospital (referred to as \textit{bpic2011}). Process remaining time prediction using the log is considered for model inspection and analysis.


\paragraph{BPIC 2012:} The event log records the execution of a loan application process at a Dutch financial institution. For process outcome prediction, each trace in the log is labelled as \textit{``accepted''}, \textit{``declined''}, or \textit{``cancelled''} (based on whether the trace contains the occurrence of activity \texttt{O\_ACCEPTED}, \texttt{O\_DECLINED}, or \texttt{O\_CANCELLED}). One of the three logs -- \textit{bpic2012\_1} which concerns loan acceptance -- is considered. 

\paragraph{BPIC 2015:} There are five event logs recording the execution of a permit application process at five Dutch municipalities, respectively. For process outcome prediction, the rule stating ``every occurrence of activity~\texttt{01\_HOOFD\_020} is eventually followed by activity~\texttt{08\_AWB45\_020\_1}'' is used to label the outcome of a trace as \textit{positive}~\citep{teinemaa2019}. 
The log for one of the municipalities --  \textit{bpic2015\_5} -- is considered. 

\begin{table}[!h]
\caption{Statistics of the three event logs used for model inspection and analysis}
\label{tab:my-table}
\resizebox{\columnwidth}{!} {
\begin{tabular}{l|cccccccc}
\hline
\multicolumn{1}{c}{\textbf{\begin{tabular}[c]{@{}c@{}} event\\ log\end{tabular}}} &
\multicolumn{1}{c}{\textbf{\begin{tabular}[c]{@{}c@{}}\# cases\\ ~ \end{tabular}}} &
\multicolumn{1}{c}{\textbf{\begin{tabular}[c]{@{}c@{}} \# events\\ ~\end{tabular}}} &
\multicolumn{1}{c}{\textbf{\begin{tabular}[c]{@{}c@{}} \# activities\\ ~\end{tabular}}} &
\multicolumn{1}{c}{\textbf{\begin{tabular}[c]{@{}c@{}}\# static\\ attributes\end{tabular}}} & \multicolumn{1}{c}{\textbf{\begin{tabular}[c]{@{}c@{}}\# dynamic\\ attributes\end{tabular}}} & \multicolumn{1}{c}{\textbf{\begin{tabular}[c]{@{}c@{}} min\\ length\end{tabular}}} & \multicolumn{1}{c}{\textbf{\begin{tabular}[c]{@{}c@{}} max \\ length\end{tabular}}} & \multicolumn{1}{c}{\textbf{\begin{tabular}[c]{@{}c@{}} avg\\ length\end{tabular}}} \\
\hline
bpic2011        & 1,140     & 149,730   & 251   & 6     & 14    & 1     & 1,814 & 25.0 \\
bpic2012\_1     & 4,685     & 186,693   & 36    & 1     & 10    & 15    & 175   & 35.0 \\
bpic2015\_5     & 1,051     & 54,562    & 217   & 18    & 12    & 5     & 134   & 50.0 \\
\hline
\end{tabular}
}
\end{table}

\paragraph{Notations:} The feature representations in both local and global explanations used in model inspection in the next section are to be read as follows: 
\begin{itemize}
    \item \texttt{agg\_\_[Activity]|[Resource]} represents the frequency of an activity executed by a resource) in a trace via aggregation encoding (\textbf{agg}).
    \item \texttt{static\_\_[attribute]} represents the case attribute for a trace (of which the value does not change during the execution of the trace).
\end{itemize}

\subsection{Analysis and Discussion}
\label{subsec:analysis} 



\subsubsection{Detecting Data Leakage} 

We inspect the business process prediction models trained for process outcome prediction using event log~\textit{bpic2015\_5}. The following two combinations of bucketing and encoding methods are applied respectively with XGBoost: i) single bucket and aggregation encoding (\textit{single\_agg}) and ii) prefix length bucket and aggregation encoding (\textit{prefix\_agg}). 

\begin{figure}[htbp]
    \includegraphics[width=.95\textwidth]{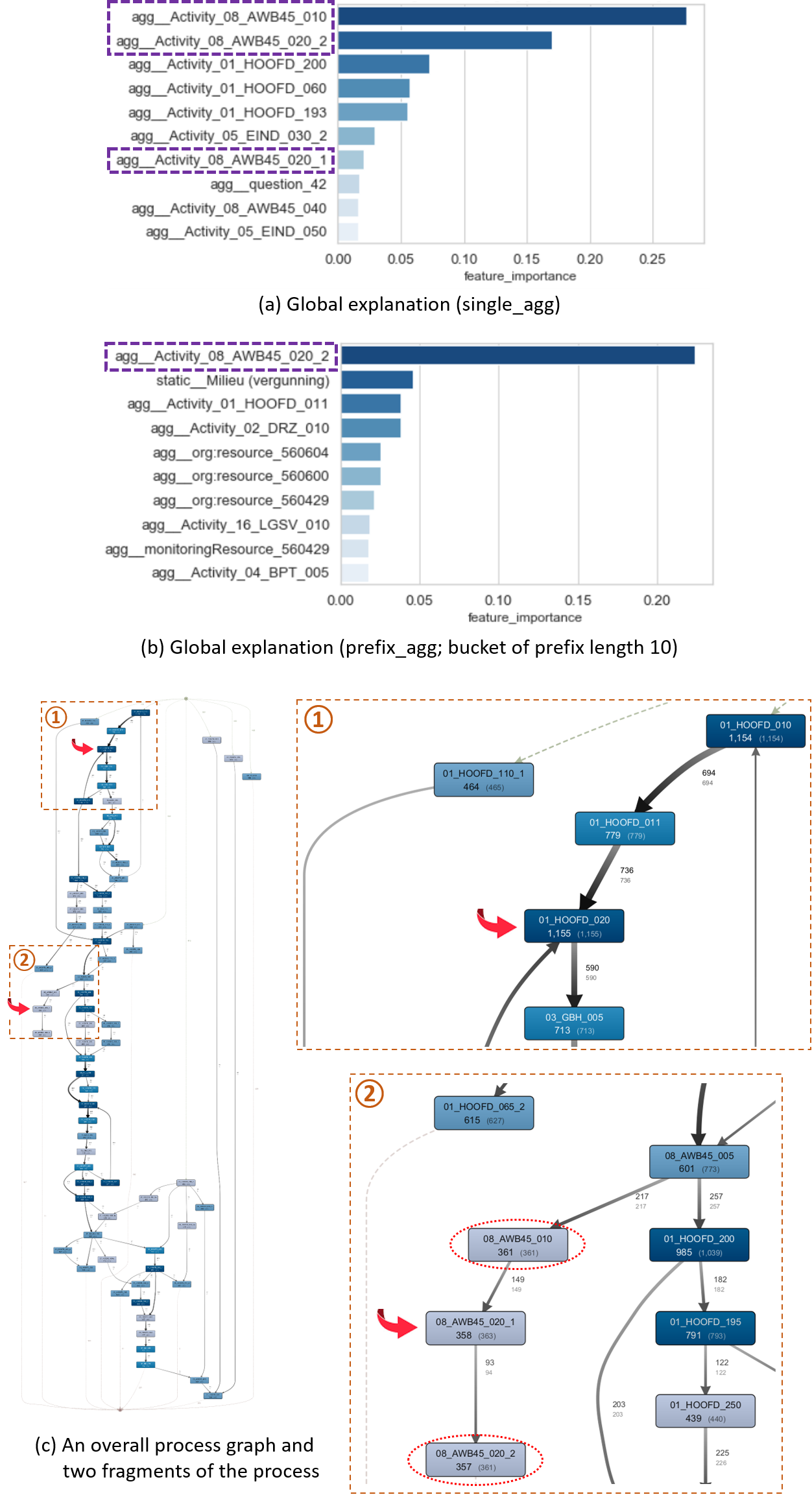}
    \caption{Inspecting process outcome prediction models for \textit{bpic2015\_5} event log with XGBoost: (a) global explanation of the model using \textit{single\_agg} method; (b) global explanation of the model trained for bucket of prefix length~$10$ using \textit{prefix\_agg} method; and (c) an overall process graph and two fragments of the process captured by the event log.}
\label{fig:bpic2015_inspect}
\end{figure}

Fig.~\ref{fig:bpic2015_inspect}(a) shows the global explanation for the model using \textit{single\_agg} method. The occurrences of three activities \texttt{08\_AWB45\_010}, \texttt{08\_AWB45\_020\_2} and \texttt{08\_AWB45\_020\_1} are identified among the important features for outcome prediction. However, the occurrence of \texttt{08\_AWB45\_020\_1} is part of the outcome to be predicted (as described in Section~\ref{subsec:data}). Fig.~\ref{fig:bpic2015_inspect}(b) shows the global explanation for the model trained for the bucket of prefix length~$10$ using \textit{prefix\_agg} method with XGBoost. The activity \texttt{08\_AWB45\_020\_2} is identified as the most important feature used for prediction. 
Fig.~\ref{fig:bpic2015_inspect}(c) depicts an overall process graph of the process execution captured by event log~\textit{bpic2015\_5} and zoom-in views of two fragments of the process: 
one contains activity~\texttt{01\_HOOFD\_020}\footnote{As described in Section~\ref{subsec:data}, the occurrence of activity \texttt{01\_HOOFD\_020} eventually to be followed by the occurrence of activity \texttt{08\_AWB45\_020\_1} is part of the rule for labelling process outcome for event log~\textit{bpic2015\_5}.}, the other involves activities~\texttt{08\_AWB45\_010}, \texttt{08\_AWB45\_020\_1} and \texttt{08\_AWB45\_020\_2}. 

For both prediction models, global explanations together with the knowledge of the business process (discovered from event log data via process mining techniques) can be used to identify potential issues with the prediction model under inspection. As shown in Fig.~\ref{fig:bpic2015_inspect}(c), each activity in the process graph is also annotated with case frequency -- the digit displayed under the activity name -- which specifies the number of cases involving the occurrence of the activity. 
A statistical analysis over the process graph indicates that: 
i) activity \texttt{08\_AWB45\_020\_2} is executed after \texttt{08\_AWB45\_020\_1} in 68\% of the cases, and 
ii) activity \texttt{08\_AWB45\_010} occurs prior to \texttt{08\_AWB45\_020\_1} in 50\% of the cases. 

All these observations reveal that the prediction model of \textit{single\_agg} with XGBoost exhibits a problem of \textbf{data leakage}~\citep{kaufman2011}, ``where information about the label of prediction that should not legitimately be available is present in the input''. As such, the features that occur along with or after the activity used as the label have likely influenced the model predictions. 


\subsubsection{Investigating Feature Relevance}

The purpose of this analysis is to reason the relevance of the features identified important for process prediction. Feature importance in global explanations and feature impact to predictions of traces in local explanations are valuable inputs to investigating feature relevance. At the same time, relevant knowledge about process execution is also needed, and often domain knowledge may also be required in order to understand how relevant a feature is to a given context (such as medical scenarios). Below, we discuss two examples of model inspection for analysing feature relevance: i) process outcome prediction using event log~\textit{bpic2012\_1} and ii) process remaining time prediction using log~\textit{bpic2011}. 

\paragraph{Analysis~I:} 

We inspect the prediction model trained for process outcome prediction of event log~\textit{bpic2012\_1} using a \textit{single\_agg} method and a single classifier XGBoost (trained for a single bucket containing all traces of all prefix lengths). 

\begin{figure}[htbp]
    \includegraphics[width=.9\textwidth]{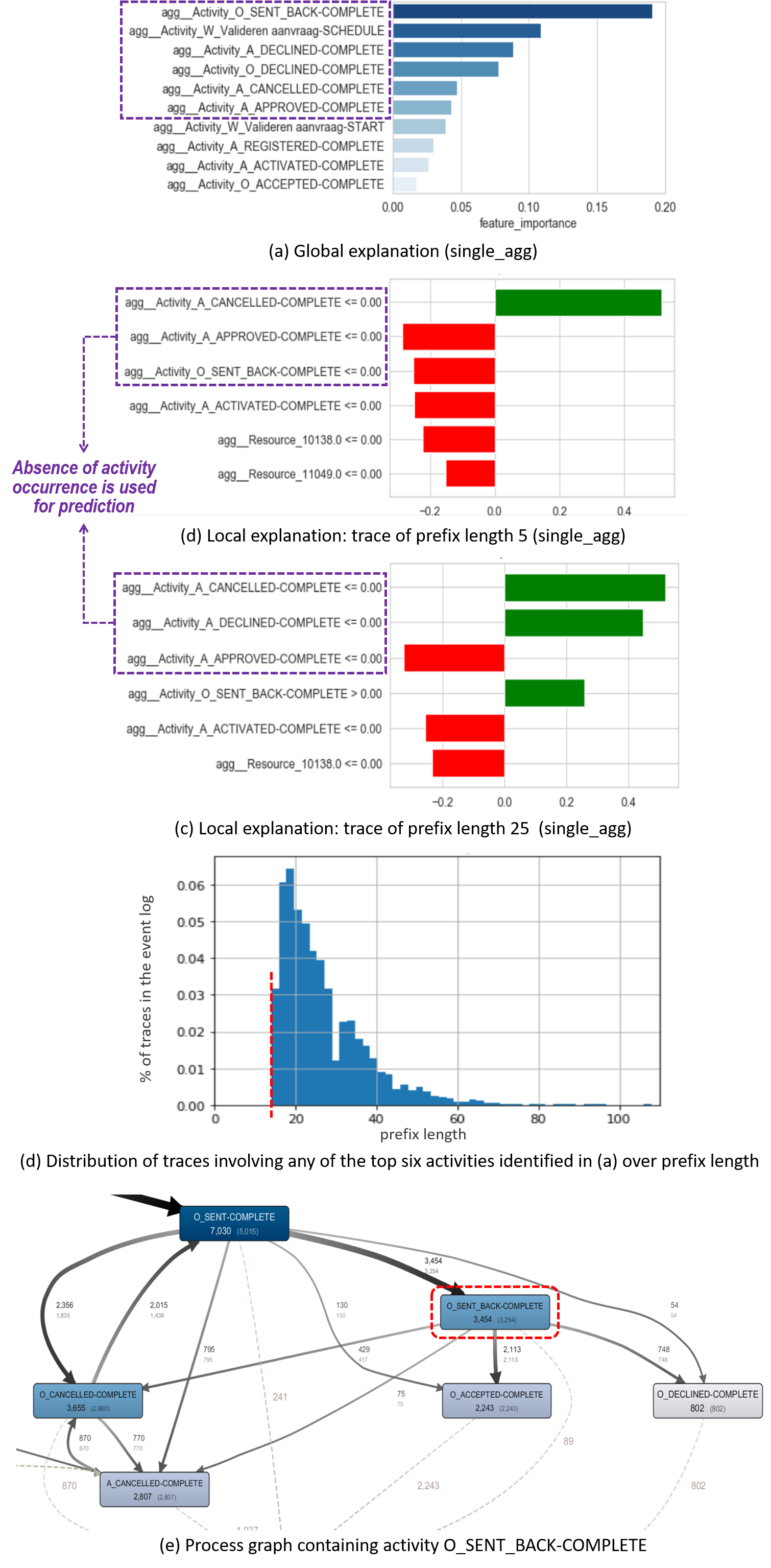}
    \caption{Inspecting process outcome prediction model for \textit{bpic2012\_1} event log using \textit{single\_agg} method with XGBoost: 
    (a) global explanation;  
    (b) and (c) local explanations for traces of prefix lengths $5$ \& $25$;  
    (d) distribution of certain trace occurrences over prefix length used for prediction; and 
    (e) fragment of discovered process containing activity \texttt{O\_SENT\_BACK}.}
\label{fig:bpic2012_inspect}
\end{figure}

Starting with global explanations, Fig.~\ref{fig:bpic2012_inspect}(a) shows the importance of the features used by the model, in which the occurrences of activities \texttt{O\_SENT\_BACK}, \texttt{W\_Validate Request}, \texttt{A\_DECLINED}, \texttt{O\_DECLINED}, \texttt{A\_CANCELLED} and \texttt{A\_APPROVED} are the top six important features\footnote{For brevity, we omit the activity label suffix `\texttt{-COMPLETED}' in the discussion.}. We then generate local explanations for two randomly chosen traces for which the model correctly predicted the positive outcome (i.e., loan accepted). Fig.~\ref{fig:bpic2012_inspect}(b) and~(c) illustrate local explanations for traces with prefix lengths of~$5$ and~$25$, respectively. In both, the first three important features are given by the \textit{absence of occurrences of specific activities}, and these activities are among the top six important features in the global explanation (Fig.~\ref{fig:bpic2012_inspect}(a)). Also, taking as an example the most important feature in both local explanations, the absence of the occurrence of activity~\texttt{A\_CANCELLED} has a significant impact on the prediction of loan being accepted, and this seems reasonable for such outcome prediction. 

We are interested in obtaining relevant knowledge about process execution captured by \textit{bpic2012\_1} event log for further investigation. A statistical analysis over the discovered process from the log reveals that the top six important activities identified in the global explanation (Fig.~\ref{fig:bpic2012_inspect}(a)) occur in the traces with a \textit{minimum} prefix length of~$14$ during process execution (refer to  Fig.~\ref{fig:bpic2012_inspect}(d)). This means that, for any trace with a prefix length less than~$14$, none of these six activities (including e.g., \texttt{A\_CANCELLED}) can occur, and hence the absence of any of these activities is \textit{irrelevant} for making a prediction for such a trace. Given this knowledge, using the absence of any of these activities as an important feature to make an outcome prediction for the trace with a prefix length of~$5$ (Fig.~\ref{fig:bpic2012_inspect}(b)) is \textit{not reliable}.


In contrast, we observe that for a trace with a (longer) prefix length of~$25$ (Fig.~\ref{fig:bpic2012_inspect}(c)), the absences of activities \texttt{A\_CANCELLED}, \texttt{A\_DECLINED} and \texttt{A\_APPROVED} can be relevant features for prediction, which is also consistent with feature importance provided in the global explanation. 
Furthermore, the model also uses the occurrence of activity \texttt{O\_SENT\_BACK} as an important feature for predicting a positive outcome, which is also relevant feature as the process discovered from the event log indicates that a loan will likely be accepted after the occurrence of \texttt{O\_SENT\_BACK} (see Fig.~\ref{fig:bpic2012_inspect}(e)). 



\paragraph{Analysis~II:}  

We inspect the prediction model trained for process remaining prediction of \textit{bpic2011} event log using a \textit{single\_agg} method with XGBoost. 

\begin{figure}[htbp]
    \includegraphics[width=.95\textwidth]{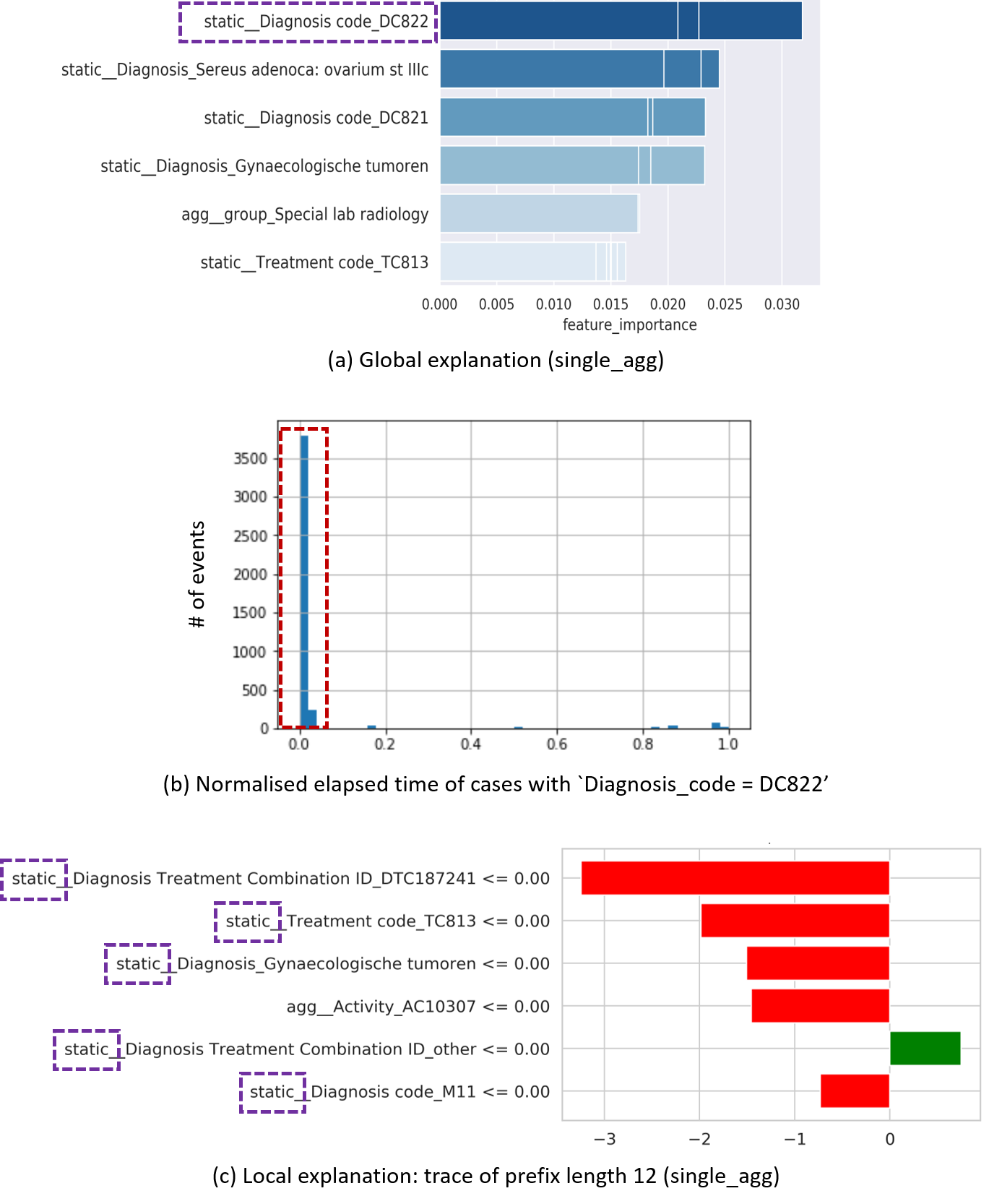}
    \caption{Inspecting process remaining time prediction for \textit{bpic2011} event log using \textit{single\_agg} with XGBoost: 
    (a) global explanation; 
    (b) normalised elapsed time for diagnosis code `DC822'; and 
    (c) local explanation for trace of prefix length~$12$.}
\label{fig:bpic2011_inspect}
\end{figure}

As an interesting observation, the global explanation shown in Fig.~\ref{fig:bpic2011_inspect}(a) indicates that the most significant feature for remaining time prediction is \texttt{Diagnosis\_code = DC822}. A statistical analysis over the cases of process execution related to this particular diagnosis code reveals that 82\% of the events associated with this diagnosis code had the feature \texttt{elapsed time = 0} (refer to  Fig.~\ref{fig:bpic2011_inspect}(b)), which means that the corresponding activity starts and ends immediately. To this end, it is necessary to gain relevant domain knowledge, in addition to process knowledge, to derive further insights about the relationship between this particular diagnosis code, the observation of 0-valued elapsed time, and their relation with the remaining time till a running case of the process is completed. 

Next, from the local explanation of a randomly chosen trace with a prefix length of~$12$ shown in Fig.~\ref{fig:bpic2011_inspect}(c), it can be observed that the prediction model uses  features like \texttt{Diagnosis Treatment Combination ID}, \texttt{Treatment code} and  \texttt{Diagnosis code} in order to determine the remaining time of the case. 
Such explanation reveals that the model is relying mostly on \textbf{static} features, which are the features that do not change throughout the lifetime of a case. 
The usage of static features suggests that i) the process execution does not rely on the dynamic aspect of process execution (such as the sequence of activities that have occurred), and ii) the model uses attributes that do not change during the case execution when making a prediction. Hence, the relevance of these static features for remaining time prediction, which is a typical regression problem, is questionable. 

\section{Conclusion}
\label{sec:concl}

We have presented an approach to support inspection of business process prediction models by leveraging both the explanations generated by interpretable machine learning mechanisms and the process knowledge extracted from event logs via process mining techniques. The approach has been used to guide the design of several experiments of model inspection in the context of two existing predictive process monitoring benchmarks and three real-life event logs. The results of model inspection have been analysed leading to interesting findings on data leakage and use of irrelevant features in existing models for business process predictions. 

From this work, we have also learned that: 
i) to gain a deeper understanding of model explanations, the domain knowledge of a particular field is also necessary in addition to relevant process knowledge that can be extracted from event logs; 
ii) relying on conventional model performance measures (e.g., accuracy, precision) is not adequate, and model reliability should be included as another important aspect for evaluation; and 
iii) results from model inspection can help with model tuning to improve the reliability and robustness of process prediction models. These will inform directions of our future work. 


\paragraph{\bf Acknowledgement:} We thank the authors of the two predictive process monitoring benchmarks~\citep{teinemaa2019,verenich2019} for the high quality code they released, which allowed us to conduct experiments of model inspection on selected business process prediction models. 



\bibliographystyle{spbasic}      
\bibliography{sections/ref}   

%
%

\end{document}